\documentclass[10pt,twocolumn,letterpaper]{article}

\usepackage{cvpr}
\usepackage{times}
\usepackage{epsfig}
\usepackage{graphicx}
\usepackage{amsmath}
\usepackage{amssymb}

\usepackage[breaklinks=true,bookmarks=false]{hyperref}

\cvprfinalcopy 

\def\cvprPaperID{32} 

\ifcvprfinal\fi

\begin{document}

\title{ Robust One Shot Audio to Video Generation}

\author{Neeraj Kumar\\
{Hike Private Limited, India}\\
{\tt\small neerajku@hike.in}
\and Srishti Goel\\
{Hike Private Limited, India}\\
{\tt\small srishtig@hike.in}
\and Ankur Narang\\
{Hike Private Limited, India}\\
{\tt\small ankur@hike.in}
\and Mujtaba  Hasan\\
{Hike Private Limited, India}\\
{\tt\small mujtaba@hike.in}
}

\maketitle

\begin{abstract}
Audio to Video generation is an interesting problem that has numerous applications across industry verticals including film making, multi-media, marketing, education and others. High-quality video generation with expressive facial movements is a challenging problem that involves complex learning steps for generative adversarial networks. Further, enabling one-shot learning for an unseen single image increases the complexity of the problem while simultaneously making it more applicable to practical scenarios.\par
	In the paper, we propose a novel approach~\textit{OneShotA2V} to synthesize a talking person video of arbitrary length using as input: an audio signal and a single unseen image of a person. OneShotA2V leverages curriculum learning to learn movements of expressive facial components and hence generates a high-quality talking head video of the given person.
\par Further, it feeds the features generated from the audio input directly into a generative adversarial network and it adapts to any given unseen selfie by applying few-shot learning with only a few output updation epochs. OneShotA2V leverages spatially adaptive normalization based multi-level generator and multiple multi-level discriminators based architecture. The input audio clip is not restricted to any specific language, which gives the method multilingual applicability. Experimental evaluation demonstrates superior performance of OneShotA2V as compared to Realistic Speech-Driven Facial Animation with GANs(RSDGAN)~\cite{Alpher05}, Speech2Vid~\cite{Chung17b}, and other approaches, on multiple quantitative metrics including: SSIM (structural similarity index), PSNR (peak signal to noise ratio) and CPBD (image sharpness). Further, qualitative evaluation and Online Turing tests demonstrate the efficacy of our approach.

\end{abstract}

\section{Introduction}
Audio to Video generation has numerous applications across industry verticals including film making, multi-media, marketing, education and others. In the film industry, it can help through automatic generation from the voice acting and also occluded parts of the face. Additionally, it can help in limited bandwidth visual communication by using audio to auto-generate the entire visual content or by filling in dropped frames. High-quality video generation with expressive facial movements is a challenging problem that involves complex learning steps. 
%
%
\par The audio or speech signal contains rich information about the mood (expression) and the intent of the user. On hearing the audio, one can predict the sentiment or emotion depicted by the user. This expressive power of the audio can be used to generate robust and high-quality talking head videos. The talking-head video aims to handle expressive facial movements and head movement based on the audio content and expression.

\par Most of the work in this field has been centered towards the mapping of audio features (MFCCs, phonemes) to visual features (Facial landmarks, visemes etc.)~\cite{Alpher04,audio2face,visemes1,visemes2}. Further computer graphics techniques select frames of a specific person from the database to generate expressive faces. Few techniques which attempt to generate the video using raw audio focuses for the reconstruction of the mouth area only~\cite{Chung17b}. Due to a complete focus on lip-syncing, the aim of capturing human expression is ignored. Further, such methods lack smooth transition between frames which does not make the final video look natural. Regardless, of which approach we use, the methods described above are either subject dependent~\cite{article12,temporalLoss} or generate unnatural videos~\cite{Wiles18}  due to lack of smooth transition and/or require high compute time to generate video for a new unseen speaker image for ensuring high-quality output~\cite{NeuralHead1}.\par

We propose a novel approach that is capable of developing a speaker-independent and language-independent high-quality natural-looking talking head video from a single unseen image and an audio clip. Our model captures the word embeddings from the audio clip using a pre-trained deepspeech2 model~\cite{DeepSPeech2} trained on Librispeech corpus~\cite{librispeech}. These embeddings and the image are then fed to the multi-level generator network which is based on the Spatially-Adaptive Normalization architecture~\cite{park2019SPADE}. Multiple multi-level discriminators~\cite{wang2018pix2pixHD} are used to ensure synchronized and realistic video generation. A multi-scale frame discriminator is used to generate high-quality realistic frames. A multi-level temporal discriminator is modeled which ensures temporal smoothening along with spatial consistency. Finally, to ensure lip synchronization we use SyncNet architecture~\cite{assael2016lipnet} based discriminator applied to the lower half of the image. To make the generator input-time independent, a sliding window approach is used. Since, the generator needs to finally learn to generate multiple facial component movements along with high video quality, multiple loss functions both adversarial and non-adversarial are used in a curriculum learning fashion. For fast low-cost adaptation to an unseen image, a few output updation epochs suffice to provide one-shot learning capability to our approach. 

Specifically, we make the following contributions:\\
$(a)$ We present a novel approach, $\bf{OneShotA2V}$, that leverages curriculum learning to simultaneously learn movements of expressive facial components and generate a high-quality talking-head video of the given person.\\ 
$(b)$ Our approach feeds the features generated from the audio input directly into a generative adversarial network and it adapts to any given unseen selfie by applying one-shot learning with only a few output updation epochs.\\
$(c)$ It leverages spatially adaptive normalization based multi-level generator and multiple multi-level discriminators based architecture to generate video which simultaneously considers lip movement synchronization and natural facial expressions incorporating eye blink and eyebrow movements along with head movement. \\
$(d)$ Experimental evaluation on the GRID~\cite{Alpher03} datasets, demonstrates superior performance of OneShotA2V as compared to Realistic Speech-Driven Facial Animation with GANs(RSDGAN)~\cite{Alpher05}, Speech2Vid~\cite{Chung17b}, and other approaches, on multiple quantitative metrics including: SSIM (structural similarity index), PSNR (peak signal to noise ratio) and CPBD (image sharpness). Further, qualitative evaluation and Online Turing tests demonstrate the efficacy of our approach.

\section{Related Work}

A lot of work has been done in synthesizing realistic videos from audio with an image as an input. Speech is a combination of content and expression and there is a perceptual variability of speech that exists in the form of various languages, dialects and accents. The understanding and modeling of the speech are more complicated as compared to the text which is devoid of various aspects of speech. \par

Various works have been done in this aspect to understand the different aspects of speech to generate realistic speech-driven videos. \par

The earliest methods for generating videos relied on Hidden Markov Models which captured the dynamics of audio and video sequences. Simons and Cox~\cite{SimonCox} used the Viterbi algorithm to calculate the most likely sequence of mouth shape given the particular utterances. Such methods are not capable of generating quality videos and lack emotions.

\subsection{Phoneme and Visemes based classification of speech}
Phoneme and Visemes based approaches have been used to generate the videos. Real-Time Lip Sync for Live 2D Animation~\cite{Alpher04} has used an LSTM based approach to generate live lip synchronization on 2D character animation.\par

Some of these methods target rigged 3D characters or meshes with predefined mouth blend shapes that correspond to speech sounds ~\cite{inproceedings1,article09,inproceedings3,article10,article11,article12}, while others generate 2D motion trajectories that can be used to deform facial images to produce continuous mouth motions~\cite{2dpaper,2dpaper1}. These methods are primarily focused on mouth motions only and do not show emotions such as eye blinking, eyebrows movements, etc.

\subsection{Video synthesis using deep networks}
CNN has been used for generating the videos by giving audio features to the network. Audio2Face~\cite{audio2face} model uses the CNN method to generate an image from audio signals. You said That~\cite{Chung17b}(Speech2Vid) has used an encoder-decoder based approach for generating realistic videos. MFCC coefficients of audio signals are being used as an input. L1 loss at the pixel level is used between synthesized image and target image which penalizes any deviation of the generated image from the target one. This disincentivizes the model to generate realistic images without spontaneous expressions except for mouth movement. Our approach uses a spatially adaptive network instead of the encoder as used in \cite{Chung17b} to learn the parameters of an image.  \par

Synthesizing Obama: Learning Lip Sync from Audio~\cite{article12} is able to generate quality videos of Obama speaking with accurate lip-sync. They use RNN based approach to map from raw audio features to mouth shapes. This method is trained on a single target image to generate high-quality videos. Our approach is able to generate videos on a single unseen image using a spatially adaptive network and a one-shot approach.
LumièreNet: Lecture Video Synthesis from Audio~\cite{temporalLoss} is generating high-quality, full-pose headshot lecture videos from the instructor’s new audio narration of any length. They have used dense pose~\cite{densepose}, LSTM , variational auto-encoder~\cite{vae}  and GANs based approach to synthesize the videos . The limitation is that they are not able to produce lip-synced video as they are only using dense pose for pose information. Our approach has used a synchronization discriminator for the generation of coherent lip-synced videos. They have used Pix2Pix~\cite{Pix2Pix} for the frame synthesis and we are using a spatially adaptive generator for frame generation which is able to generate higher quality videos.

The recent introduction of GANs in~\cite{Authors14b} has shifted the focus of the machine learning community to generative modeling. The generator’s goal is to produce realistic samples and the discriminator’s goal is to distinguish between the real and generated samples. However, GANs are not limited to these applications and can be extended to handle videos~\cite{wang2018vid2vid}.\par

Temporal Gan~\cite{Temporalgan} and Generating Videos with Scene Dynamics~\cite{SceneDynamics} have done the straight forward adaptation of GANs for generating videos by replacing 2D convolution layers with 3D convolution layers. Such methods are able to capture temporal dependencies but require constant length videos. Our approach is able to produce lower word error rate and generate consistent videos of variable length using multi-scale temporal discriminator and synchronization discriminator.\par

Realistic Speech-Driven Facial Animation with GANs(RSDGAN)~\cite{Alpher05} used GAN based approach to produce quality videos. They have used identity encoder, context encoder and frame decoder to generate images and used various discriminators to take care of different aspects of video generation. They have used frame discriminator to distinguish real and fake images, sequence discriminator to distinguish real and fake videos and synchronization discriminator for better lip synchronization in videos. We introduce spatially adaptive normalization along with a one-shot approach and implemented curriculum learning to produce better results. This is explained in Sections 3,4,5.~\par

Few-Shot Adversarial Learning of Realistic Neural Talking Head Models~\cite{NeuralHead1} have used meta-learning for generating the videos on unseen images taking video as an input. They have used content loss measures the distance between the ground truth image and the reconstruction using the perceptual similarity measure~\cite{PerceptualLoss} from VGG19~\cite{VGG19} network trained for ILSVRC classification and VGGFace~\cite{VGGFace} and adversarial loss. For unseen images, the model needs to run for 75 epochs on an unseen image to give better video quality. This is computationally heavy, so we are using a one-shot approach using perceptual loss during inference. Due to the spatially adaptive nature of our generator architecture, we are able to generate good quality video at a low computational cost. Few shot Video to Video Synthesis~\cite{wang2019fewshotvid2vid} is able to generate videos on unseen images given a video as an input by using a network weight generation module for extracting the pattern. Such a method is computationally heavy concerning our approach which is one shot approach in video generation.

X2face~\cite{Wiles18} model uses GANs based approach to generate videos given a driving audio or driving video and a source image as an input. The model learns the face embeddings of source frame and driving vectors of driving frames or audio basis which generates the videos. This model is trained on 1fps which can lead to un-natural video synthesis. They have used L1 loss and identity loss for video generation and have not used any loss for temporal coherency. Our approach generates the videos at 25fps which are capable of generating natural realistic videos and have incorporated multi-scale temporal discriminator and lip-sync discriminator for temporal coherency. 

The MoCoGAN~\cite{Tulyakov:2018:MoCoGAN} uses RNN based generators with separate latent spaces for motion and content. A sliding window approach is used so that the discriminator can handle variable-length sequences. This model is trained to generate disentangled content and motion vectors such that they can generate audios with different emotions and contents. Our approach has used deep speech2 features to learn content embeddings such that it is able to produce better videos with low word error rate.

Animating Face using Disentangled Audio Representations~\cite{disentangle} has generated the disentangled representation of content and emotion features to generate realistic videos. They have used variational autoencoders~\cite{vae} to learn representation and feed them into GANs based model to generate videos.  Our approach has used deep speech2 features to learn content embeddings instead of variational autoencoders. Instead of their Unet~\cite{Unet} architecture, we are using a spatially adaptive generator to generate high-quality videos.

Audio-driven Facial Reenactment~\cite{neuralhead} used AudioexpressionNet to generate 3D face model. The estimated 3D face model is rendered using the rigid pose observed from the original target image. Our approach is using a spatially adaptive and one-shot approach to generate 2D videos and are not generating 3D mesh.

Existing methodologies have worked on video generations with lip movement and expressions. Our goal is to create high-quality videos using a one-shot approach, so that we can experience high definition videos along with expressions and multilingual support.

\section{Architectural Design}

\begin{figure}
  \includegraphics[width=\linewidth]{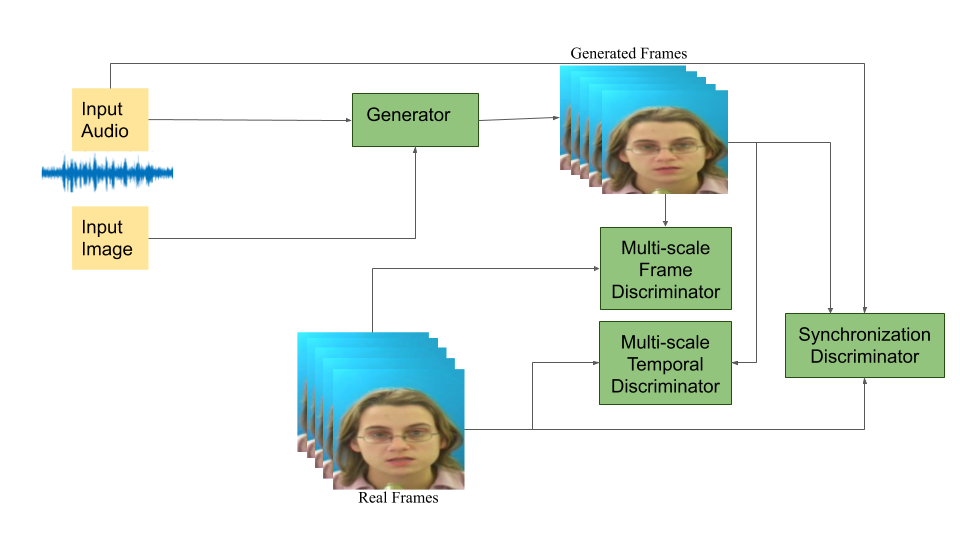}
  \caption{Model for generating robust and high-quality videos. This uses deep speech audio features to be fed into SPADE Generator and 2 discriminators i.e frame discriminator which is a multi-scale discriminator for frame generation and another discriminator for better lip synchronization.}
  \label{fig:arch}
\end{figure}

OneShotA2V consists of a single generator and 3 discriminators as shown in Figure ~\ref{fig:arch}. Each of the discriminators are used for specific purposes. Different losses are used to make the generator learn better distribution for generating realistic videos.

\begin{figure}
  \includegraphics[width=\linewidth]{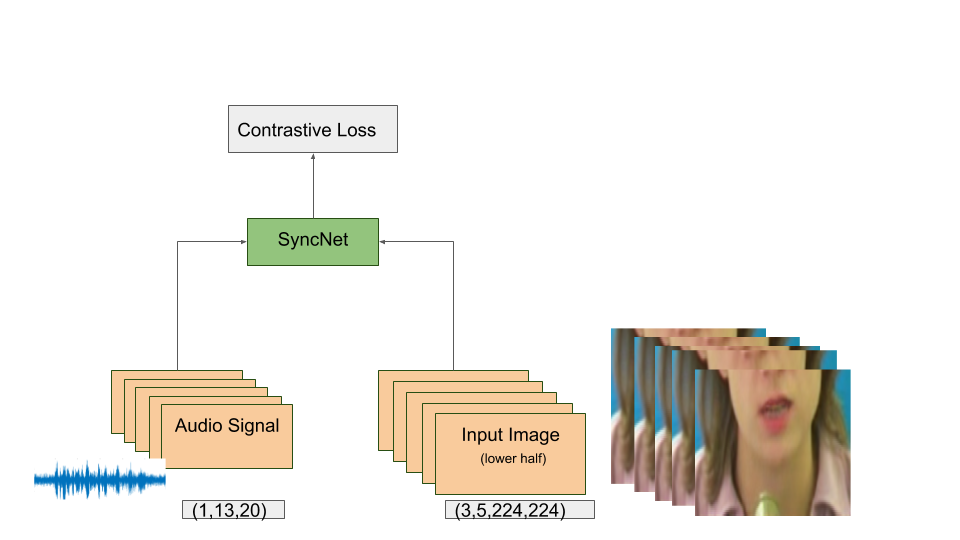}
  \caption{We have used the SyncNet architecture for better lip synchronization which is trained on GRID dataset with contrastive loss and then used its loss in our proposed architecture}
  \label{fig:syncnet}
\end{figure}

\subsection{Generator}

 The initial layers of generator, G uses deepspeech2~\cite{DeepSPeech2} layers followed by Spatially-Adaptive normalization similar to SPADE architecture ~\cite{park2019SPADE}. Conditional input of an audio frame and image is fed to the spatially adaptive network. Instead of giving the semantic input to the network as proposed in SPADE architecture, we give aligned images in an upsampling manner. This helps in the prevention of loss of information due to normalization.
 
 An audio input of 200 ms is given along with the image to produce a single frame of the video. The audio input is overlapping with the previous audio input with an overlapping interval of 0.16 ms. Every audio frame is centered around a single video frame. To do that, zero padding is done before and after the audio signal and use the following formula for the stride.
 \begin{align*}
      stride = \frac{\text{audio sampling rate}}{\text{video frames per sec}}
\end{align*}

\subsubsection{Audio features using deepspeech2 model}

The MFCC coefficients of audio input is fed into the pre-trained deepspeech2 for extracting the content-related features of audio. We have taken the few layers of deepspeech2 network and fed it as an input to the generator. This helps in improving the lip synchronization aspect for the video generation. 

\subsection{Discriminator}
We have used 3 discriminators namely a multi-scale frame discriminator, a multi-scale temporal discriminator and a synchronization discriminator.

\subsubsection{Multi-scale Frame Discriminator}

 Multi-scale discriminator~\cite{wang2018pix2pixHD}, D is used in the proposed model to distinguish the coarser and finer details between real and fake images. Adversarial training with the discriminator helps in generating realistic frames. To have high resolution generated frames, we need to have an architecture with better receptive field.  A deeper network can cause overfitting, to avoid that, multi-scale discriminators are used. Multi-scale frame discriminator consists of 3 discriminators that have an identical network structure but operate at different image scales. These discriminators are referred to as D1, D2 and D3. Specifically, we downsample the real and synthesized high-resolution images by a factor of 2 and 4 to create an image pyramid of 3 scales. The discriminators D1, D2 and D3 are then trained to differentiate real and synthesized images at the 3 different scales, respectively. The discriminators operate from coarse to fine level and help the generator to produce high-quality images.

\subsubsection{Multi-scale Temporal Discriminator}

Every frame in a video is dependent on its previous frames. To capture the temporal property along with a spatial one, we have used a multi-scale temporal discriminator~\cite{temporalLoss}. This discriminator is modeled to ensure a smooth transition between consecutive frames and achieve a natural-looking video sequence. The multi-scale temporal discriminator is described as

\begin{align*}
    L(T,G,D) &= \sum_{i=t-L}^{t}[\log(D(x\textsubscript{i}))] + [\log(D(1-G(z\textsubscript{i})))] 
\end{align*}

where t is the time instance of an audio and L is the length of the time interval for which the adversarial loss is computed.

\subsubsection{Synchronization Discriminator}

To have coherent lip synchronization, the proposed model uses SyncNet architecture proposed in Lip Sync in the wild~\cite{Chung16a}. As shown in Figure ~\ref{fig:syncnet} the input to the discriminator is an audio signal of 200ms time interval(5 audio signals of 40ms each) and 5 frames of the video. The lower half of the frame of resized to (224,224,3) is fed as an input.

\section{Curriculum Learning}

We have trained OneShotA2V in multiple phases so that it can produce better results. In the first phase we have used a multi-scale frame discriminator and applied the adversarial loss, feature matching loss and perceptual loss to learn the higher-level features of the image. When these losses stabilize, we move to the second phase in which we have added a multi-scale temporal discriminator and synchronization discriminator and used reconstruction loss, Contrastive loss and temporal adversarial loss to get a better quality image near mouth region and coherent lip synchronized high-quality videos. After the stabilization of the above losses, we have added blink loss in the third phase to generate a more realistic image capturing emotions such as eye movement and eye blinks. 

\subsection {Losses}

OneShotA2V is trained with different losses to generate realistic videos as explained below.

\subsubsection{Adversarial Loss}
Adversarial Loss is used to train the model to handle adversarial attacks and ensure generation of high-quality images for the video. The loss is defined as:

\begin{align*}
    L\textsubscript{GAN}(G,D) &= E\textsubscript{x$\sim$P\textsubscript{d}}[\log(D(x))] + E\textsubscript{z$\sim$P\textsubscript{z}}[\log(D(1-G(z)))] 
\end{align*}

 where G tries to minimize this objective against an adversarial D that tries to maximize.

\subsubsection{Reconstruction loss}
Reconstruction loss~\cite{RLoss} is used on the lower half of the image to improve the reconstruction in mouth area. L1 loss is used for this purpose as described below:

\begin{align*}
    L\textsubscript{RL} &= \sum_{n\epsilon [0,W]*[H/2,H]}^{}(R\textsubscript{n} - G\textsubscript{n})
\end{align*}

where, R\textsubscript{n} and G\textsubscript{n} are the real and generated frames respectively.

\subsubsection{Feature Loss}

Feature-matching Loss~\cite{wang2018pix2pixHD} ensures generation of natural-looking  high-quality frames. We take the L1 loss of between generated images and real images for different scale discriminators and then sum it all. We extract features from multiple layers of the discriminator and learn to match these intermediate representations from the real and the synthesized image. This helps in stabilizing the training of the generator. The feature matching loss,   L\textsubscript{FM}(G,D\textsubscript{k}) is given by:

\begin{align*}
    L\textsubscript{FM}(G,D\textsubscript{k}) &= E\textsubscript{(x,z)} \sum_ {n=1}^{T}[\frac{1}{N\textsubscript{i}}||D\textsubscript{k}^{(i)}(x)-D\textsubscript{k}^{(i)}(G(z))||\textsubscript{1}]
\end{align*}

where, T is the total number of layers and N\textsubscript{i} denotes the
number of elements in each layer.

\subsubsection{Perceptual Loss}
The perceptual similarity metric is calculated between the generated frame and the real frame. This is done by using features of a VGG19~\cite{VGG19} model trained for ILSVRC classification and VGGFace~\cite{VGGFace} dataset.The perceptual loss~\cite{PerceptualLoss},(L\textsubscript{PL}) is defined as:

\begin{align*}
    L\textsubscript{PL} &= \lambda\sum_{n=1}^{N}[\frac{1}{M\textsubscript{i}}||F^{(i)}(x)-F^{(i)}(G(z))||\textsubscript{1}]
\end{align*}
where, $\lambda$ is the weight for perceptual loss and $F^{(i)}$ is the ith layer of VGG19 network with M\textsubscript{i} elements of VGG layer.

\subsubsection{Contrastive Loss}
For coherent lip synchronization, we use the Synchronization Discriminator with Contrastive loss. The training objective is that the output of the audio and the video networks are similar for genuine pairs, and different for false pairs.

Contrastive loss,(L\textsubscript{CL}) is given by following equation 
\begin{align*}
    L\textsubscript{CL} &= \frac{1}{2N}\sum_{n=1}^{N}(y\textsubscript{n})d^2\textsubscript{n}+(1-y\textsubscript{n})max(margin-d\textsubscript{n},0)^2 &
\end{align*}

\begin{align*}
    d\textsubscript{n} &= ||v\textsubscript{n}-a\textsubscript{n}||\textsubscript{2}
\end{align*}

where, v\textsubscript{n} and a\textsubscript{n}  are fc\textsubscript{7} vectors for video and audio inputs respectively. y $\epsilon$ [0,1] is the binary similarity metric for video and audio input.

\subsubsection{Blink loss }

\begin{figure}[h!]
  \begin{center}
   \includegraphics[width=0.6\linewidth]{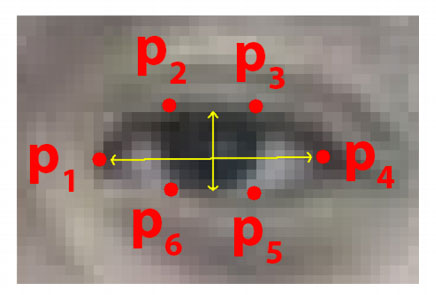}
   \caption{Spatio-Temporal Normalization Architecture}
   \label{fig:blink}
  \end{center}
\end{figure}

We have used the eye aspect ratio (EAR) taken from Real-Time Eye Blink Detection using Facial Landmarks ~\cite{Authors14} to calculate the blink loss. A blink is detected at the location where a sharp drop occurs in the EAR signal. Loss is defined as:

\begin{align*}
  m &= \frac{||p2-p6|| + ||p3-p5||}{||p1-p4||}
\end{align*}

\begin{align*}
    L\textsubscript{BL} &= ||m\textsubscript{r}-m\textsubscript{g}||
\end{align*}

where, p\textsubscript{i} is described in Figure ~\ref{fig:blink}. We have taken the L1 loss of eye aspect ratio(EAR) between real image m\textsubscript{r}  and synthesized frame m\textsubscript{g}.

\section{Few shot learning}

To achieve a more sharp and a better image quality for an unseen subject, we have used one shot approach using perceptual loss during inference time. Our approach is computationally less expensive as compared to ~\cite{NeuralHead1,wang2019fewshotvid2vid} which we have described in Section 2 and because of the spatially adaptive nature of generator architecture, we are able to achieve high-quality video. We run the model for 5 epochs during inference time to get high-quality video frames.

\section{Experiments and Results}

\subsection{Datasets and Training}
We have used the GRID dataset~\cite{Alpher03} and LOMBARD GRID~\cite{gridlombard} for the experiment and evaluation of different metrics. GRID dataset is a large multi-talker audiovisual sentence corpus. This corpus consists of high-quality audio and video $($facial$)$ recordings of 1000 sentences spoken by each of 34 talkers $($18 male, 16 female$)$. LOMBARD GRID dataset is a bi-view audiovisual Lombard speech corpus that can be used to support joint computational-behavioral studies in speech perception. The corpus includes 54 talkers, with 100 utterances per talker $($50 Lombard and 50 plain utterances$)$. It consists of 5400 videos generated on  54 talkers comprising 30 female talkers and 24 male talkers. 

Our model is implemented in Pytorch and takes approximately 4 days to run on 4 Nvidia V100 GPUs for training. Around 5000 and 1200 videos of the GRID dataset are used for training and testing purposes respectively. We have taken  3000  and 600 videos of the LOMBARD GRID dataset for training and testing purposes. The frames are extracted at 25fps. We have taken 16khz as sampling frequency for audio signals and used 13MFCC coefficients for 0.2 sec of overlapping audio for experimentation.

 The aligned face is generated for every speaker using facial landmark detector~\cite{inproceedings} and HopeNet~\cite{Ruiz_2018_CVPR_Workshops} for calculating the yaw, pitch and roll angles to get the most aligned faces for every speaker as an input.
 
 We take the Adam optimizer~\cite{Adam} with learning rate = 0.002 and $\beta_1$= 0.0 and $\beta_2$ = 0.90 for the generator and discriminators. The learning rate of the generator and discriminator is constant for 50 epochs and after that it decays to zeros in the next 100 epochs.

\subsection{Metrics}
{\bf 1. PSNR- Peak Signal to Noise Ratio: }It computes the peak signal to noise ratio between two images. The higher the PSNR the better the quality of the reconstructed image.

{\bf 2. SSIM- Structural Similarity Index: }It is a perceptual metric that quantifies image quality degradation.\ The larger the value the better the quality of the reconstructed image.

{\bf 3. CPBD- Cumulative Probability Blur Detection: }It is a perceptual based no-reference objective image sharpness metric based on the cumulative probability of blur detection developed at the Image.

{\bf 4. WER- Word error rate: }It is a metric to evaluate the performance of speech recognition in a given video. We have used LipNet architecture~\cite{assael2016lipnet} which is pre-trained on the GRID dataset for evaluating the WER. On the GRID dataset, Lipnet achieves 95.2 percent accuracy which surpasses the experienced human lipreaders.

{\bf 5. ACD- Average Content Distance(~\cite{Tulyakov:2018:MoCoGAN}): }It is used for the identification of speaker from the generated frames using OpenPose~\cite{cao2018openpose}. We have calculated the Cosine distance and Euclidean distance of representation of the generated image and the actual image from Openpose. The distance threshold for the OpenPose model should be 0.02 for Cosine distance and 0.20 for Euclidean distance ~\cite{acd2}. The lesser the distances the more similar the generated and actual images.

\subsection{Qualitative Results}

OneShotA2V is able to produce natural-looking high-quality videos of previously unseen input image and audio signals. The videos are able to do lip synchronization on the sentences provided to them. Videos were generated targeting different languages ensuring the proposed method is language independent and can generate videos for any linguistic community. \\

\begin{figure}[h!]
  \includegraphics[width=0.5\textwidth]{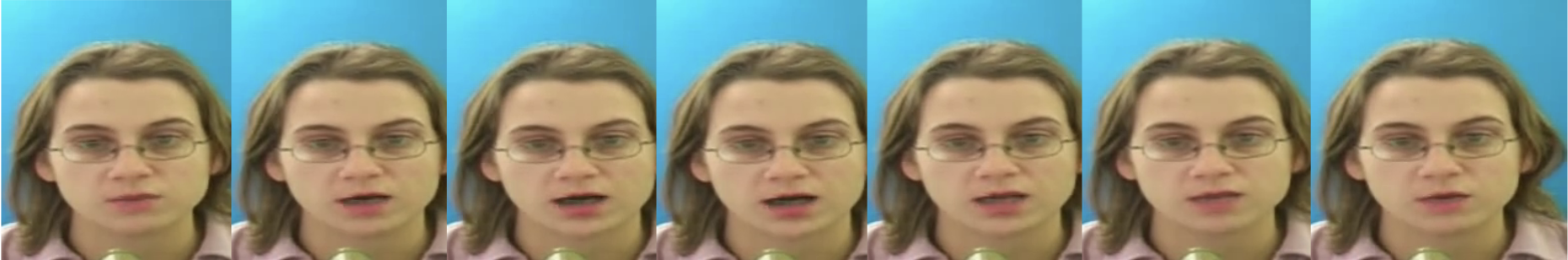}
  \caption{Female uttering the word "now"}
  \label{fig:result1}
\end{figure}

\begin{figure}[h!]
    \includegraphics[width=0.5\textwidth]{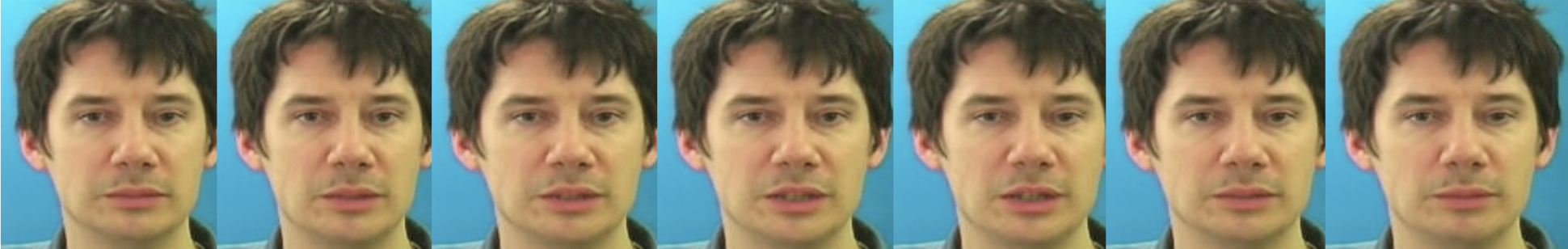}%
    \caption{Male uttering the word "bin"}%
    \label{fig:result2}
\end{figure}

Figure ~\ref{fig:result1} and Figure ~\ref{fig:result2} show different examples of the generated and lip synchronized videos for male and female test cases for the same audio clip and their ground truth frames. As observed the opening and closing of the mouth is in sync with the audio signals. Our method is able to produce synchronized lip movements displaying facial expressions such as forehead lines and eye blinks ensuring a natural-looking aesthetic output. 

\begin{figure}[h!]
    \includegraphics[width=0.5\textwidth]{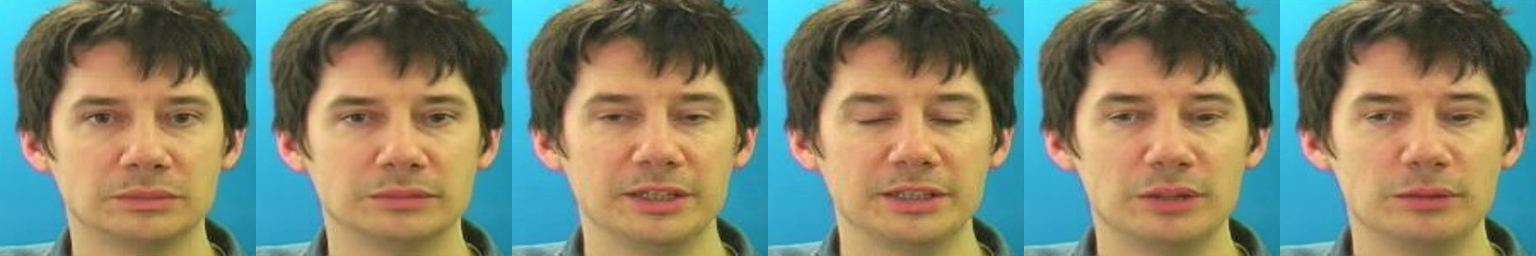}%
    \caption {Movement of eyes while speaking}%
    \label{fig:eye}
\end{figure}

Figure ~\ref{fig:eye} show the the movement of eyes of speaker while speaking . Such frames are able to generate natural videos capturing the eye's movement while speaking.\
s
\begin{figure}[h!]
    \includegraphics[width=0.5\textwidth]{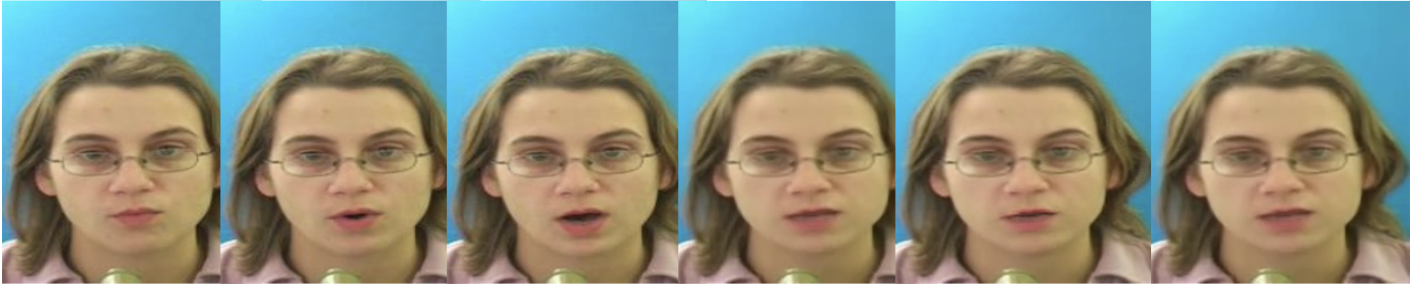}%
    \caption{Speaker uttering a hindi male name "Modi"}%
    \label{fig:modi}
\end{figure}

Figure ~\ref{fig:modi} show the generated output for a hindi audio clip ("Modi"). As observed, the generated frames are able to produce the expected lip movements and provide multilingual support.

\begin{figure}[h!]
    \includegraphics[width=0.5\textwidth]{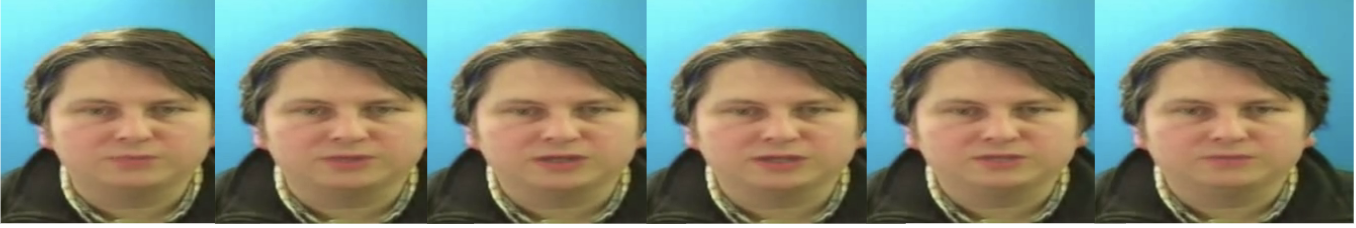}%
    \caption{Speaker uttering the word "Please" on the GRID dataset}%
    \label{fig:please}
\end{figure}

\begin{figure}[h!]
    \includegraphics[width=0.5\textwidth]{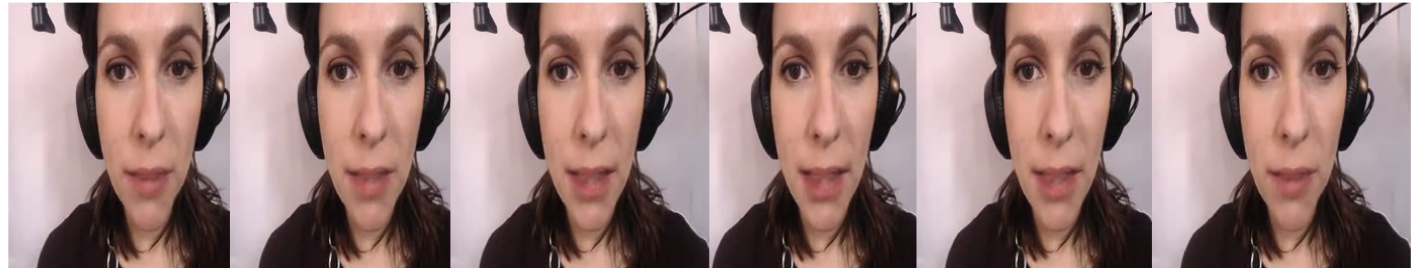}%
    \caption{Speaker uttering the word "Please" on the LOMBARD GRID dataset}%
    \label{fig:please1}
\end{figure}

Figure ~\ref{fig:please} show the generated output with the model trained on the GRID dataset and Figure ~\ref{fig:please1} show the generated output with the model trained on the LOMBARD GRID dataset.

For video clips, see the supplementary data.

\subsection{Quantitative Results}

The Proposed Model has performed better on image reconstruction metrics such as peak signal to noise ratio(PSNR) and Structural Similarity Index(SSIM) as compared to Realistic Speech-Driven Facial Animation with GANs(RSDGAN)~\cite{Alpher05} and Speech2Vid~\cite{Chung17b} Model as shown in Table ~\ref{tab:table2}. We also display the comparison with OneShotA2V trained on the LOMBARD GRID dataset ~\cite{gridlombard}. This is achieved with the use of spatially adaptive normalization in the generator architecture along with training of the proposed model in curriculum learning fashion.

\begin{table}[h!]
  \begin{center}

    \begin{tabular}{c|c|c|c} 
      \textbf{Method} & \textbf{SSIM} & \textbf{PSNR} & \textbf{CPBD}\\
      \hline
       OneShotA2V& \textbf{0.881} & \textbf{28.571} & 0.262  \\
       OneShotA2V(lombard) & 0.922 & 28.978 & 0.453 \\
       RSDGAN  & 0.818 & 27.100 & \textbf{0.268}\\
       Speech2Vid    & 0.720 & 22.662 & 0.255 \\
    \end{tabular}
    \vspace {0.25\baselineskip}
    \caption{Comparision of OneShotA2V with RSDGAN and Speech2Vid for SSIM, PSNR and CPBD}
    \label{tab:table2}
    
  \end{center}
\end{table}

\begin{table}[h!]
  \begin{center}

    \begin{tabular}{c|c|c|c} 
      \textbf{Method} & \textbf{WER} & \textbf{ACD-C} & \textbf{ACD-E}\\
      \hline
       OneShotA2V& 27.5 & 0.005 & 0.09  \\
       OneShotA2V(lombard)&26.1 &0.002 &0.064 \\
       RSDGAN  & 23.1 & - & 1.47x10\textsuperscript{-4} \\
       Speech2Vid    & 58.2 & - & 1.48x10\textsuperscript{-4}\\
    \end{tabular}
    \vspace {0.25\baselineskip}
    \caption{The above comparison is on lip synchronizing metric i.e word error rate(WER) and average content distance(ACD) by calculating cosine distance(ACD-C) and euclidean distance(ACD-E) between the actual image and the generated image. }
    \label{tab:table3}
    
  \end{center}
\end{table}

Table ~\ref{tab:table3} shows the comparison of OneShotA2V with the RSDGAN and Speech2Vid~\cite{Chung17b} models against the metrics such as word error rate (WER) to see the lip synchronizing performance of the generated videos. For this, we have used the pre-trained LipNet model whose accuracy is 95.2\% on GRID datasets. We find that OneShotA2V performed better than Speech2Vid but lagged behind RSDGAN. We have used the pre-trained OpenFace model to calculate the Cosine distance and Euclidean distance for average content distance. Experiments on OpenFace show that the distance threshold for the model should be 0.02 for cosine distance and 0.20 for euclidean distance~\cite{acd2}. 

\subsection{Psychophysical assessment}

Results are visually rated (on a scale of 5) individually by 25 persons, on three aspects,
 lip synchronization, eye blinks and eyebrow raises and quality of video.
 The subjects were shown anonymous videos at the same time for the different audio clips for side-by-side comparison. Table ~\ref{tab:table1} clearly shows that OneShotA2V performs significantly better in quality and lip synchronization which is of prime importance in videos. 


\begin{table}[h!]
  \begin{center}

    \begin{tabular}{c|c|c|c} 
      \textbf{Method} & \textbf{Lip-Sync} & \textbf{Eye-blink} & \textbf{Quality}\\
      \hline
       OneShotA2V& 90.8 & 88.5 & \textbf{76.2}  \\
       RSDGAN  & \textbf{92.8} & \textbf{90.2} & 74.3\\
       Speech2Vid    & 90.7 & 87.7 & 72.2 \\
    \end{tabular}
    \vspace {0.25\baselineskip}
    \caption{Psychophysical Evaluation (in percentages) based on users rating}
    \label{tab:table1}
  \end{center}
\end{table}

\begin{figure}[h!]
  \includegraphics[width=\linewidth]{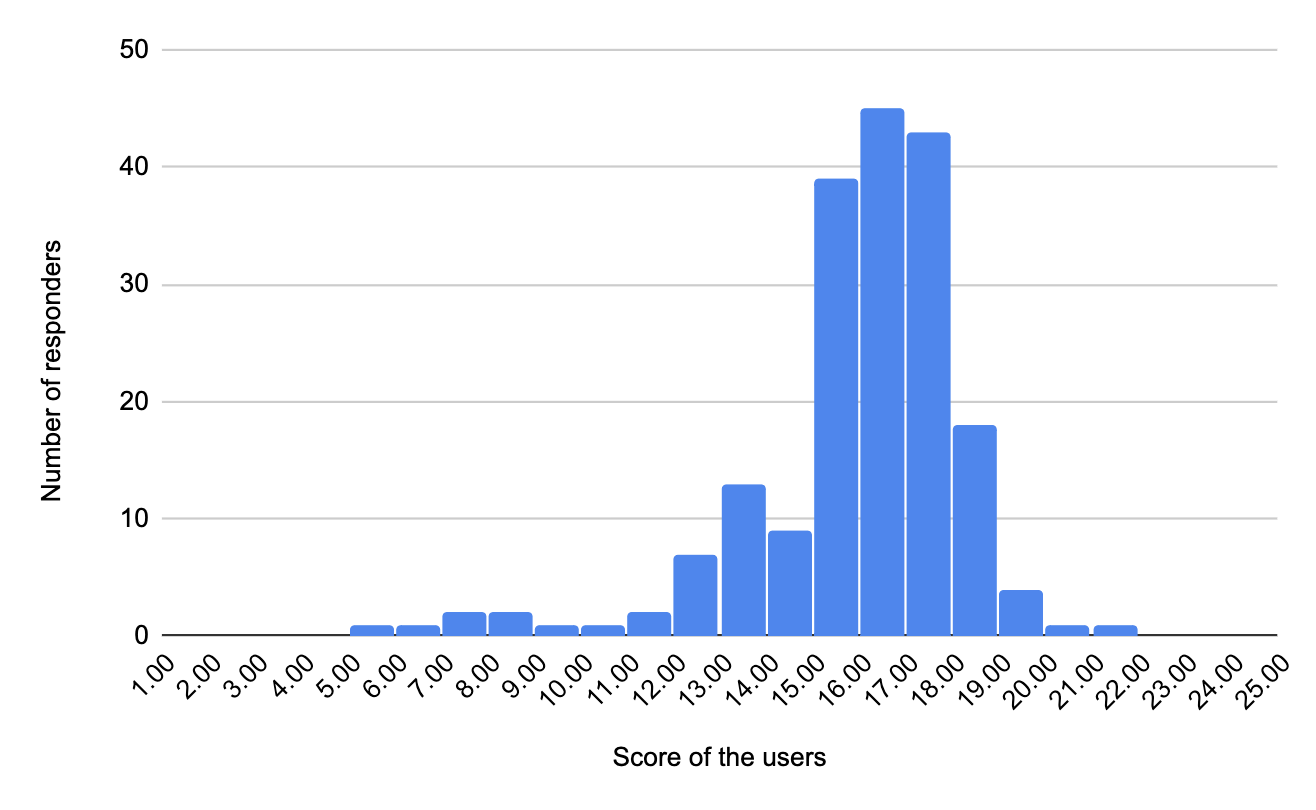}
  \caption{Distribution of user scores for the online Turing test}
  \label{fig:turing}
\end{figure}

To test the naturalism of the generated videos we conduct an online Turing test \footnote{\url{https://forms.gle/JEk1u5ahc9gny7528}}. Each test consists of 25 questions with 13 fake and 12 real videos. The user is asked to label a video real or fake based on the aesthetics and naturalism of the video. Approximately 300 user data is collected and their score of the ability to spot fake video is displayed in Figure ~\ref{fig:turing}.

\subsection{Ablation Study}

We studied the incremental impact of various loss functions on the LOMBARD GRID dataset and the GRID dataset. We have provided corresponding videos in supplementary data for better visual understanding. As mentioned in section 4 (Curriculum learning) each loss has a different impact on the final output video. Table ~\ref{tab:table4} and Table ~\ref{tab:table5}  depicts the impact of different losses on both datasets. The base model mentioned is the includes the adversarial gan loss, feature loss and perceptual loss. The addition of contrastive loss and multi-scale temporal adversarial loss in sequence discriminator helps in achieving coherent lip synchronized videos and improves the SSIM, PSNR and CPBD values. Further addition of Blink Loss, ensures improved quality of the final video. 

\begin{table}[h!]
  \begin{center}

    \begin{tabular}{c|c|c|c} 
      \textbf{Method} & \textbf{SSIM} & \textbf{PSNR} & \textbf{CPBD}\\
      \hline
       Base Model(BM)  & 0.869 & 27.996 &  0.213 \\
       BM + CL +TAL        & 0.873 & 28.327 &  0.258 \\
       BM + CL + TAL+ BL    & 0.881 & 28.571 & 0.262  \\
    \end{tabular}
    \vspace {0.25\baselineskip}
    \caption{Ablation Study on the GRID dataset where, CL is the contrastive loss ,TAL is the multi-scale temporal adversarial loss and BL is the Blink loss}
    \label{tab:table4}
    
  \end{center}
\end{table}

\begin{table}[h!]
  \begin{center}

    \begin{tabular}{c|c|c|c} 
      \textbf{Method} & \textbf{SSIM} & \textbf{PSNR} & \textbf{CPBD}\\
      \hline
       Base Model(BM)  & 0.909 & 28.656 &  0.386 \\
       BM + CL + TAL        & 0.913 & 28.712 &  0.390 \\
       BM + CL + TAL+ BL    & 0.922 & 28.978 & 0.453  \\
    \end{tabular}
    \vspace {0.25\baselineskip}
    \caption{Ablation Study on the LOMBARD GRID dataset where, CL is the contrastive loss, TAL is the multi-scale temporal adversarial loss and BL is the Blink loss}
    \label{tab:table5}
  \end{center}
\end{table}

The use of deeepspeech2 to generate audio to content embeddings helped in the improvement of WER and help us reach almost similar performance as RSDGAN.

\subsection{Conclusions and Future Work}

In this paper, we have considered robust one-shot video generation from audio input. Our approach, OneShotA2V, uses multi-level generator and multiple multi-level discriminators along with curriculum learning and few-shot learning to generate high-quality videos. Spatially adaptive normalization helps to ensure light generator architecture without encoders and also efficient few-shot learning with few updation epochs on the generator. The coherent lip movement and lower word error rate(WER) is attributed to the use of multi-scale temporal discriminator and synchronization discriminator. The use of deep speech features helps the model to learn the content vectors of audio in a better manner which led to lower word error rate.\par
Experimental evaluation on GRID dataset demonstrates superior performance of OneShotA2V as compared to Realistic Speech-Driven Facial Animation with GANs(RSDGAN)~\cite{Alpher05}, Speech2Vid~\cite{Chung17b}, and other approaches, on multiple quantitative metrics including: SSIM (structural similarity index), PSNR (peak signal to noise ratio) and CPBD (image sharpness). Further, qualitative evaluation and Online Turing tests demonstrate the efficacy of our approach. Moreover, OneShotA2V is able to perform generation of robust high-quality natural-looking videos across multiple languages without any additional requirement of multilingual datasets (audio signals).\par
In the future, we plan to add emotions so that the generated videos can capture varying degrees of emotional expressions of the speaker. This will make the video output from our approach more realistic and robust. Further, we plan to consider sophisticated curriculum learning techniques to enable the generation of more dynamic talking videos.


\end{document}